%% file: main.tex
\pdfoutput=1

\documentclass[11pt,a4paper]{article}

\usepackage [table]{xcolor}

\usepackage[hyperref]{emnlp2020}
\usepackage{times}
\usepackage{latexsym}
\usepackage{url}
\usepackage{CJKutf8}
\usepackage{xspace,mfirstuc,tabulary}
\usepackage{amsmath,bm}
\usepackage{multirow}
\usepackage{color}
\usepackage{multicol}
\usepackage{tabularx}
\usepackage{booktabs}
\usepackage{pbox}
\usepackage{framed}
\usepackage{array}
\usepackage{enumitem}
\usepackage{chngpage}
\usepackage{threeparttable}
\usepackage{dsfont}
\usepackage{graphicx}
\usepackage{enumitem}
\usepackage{epigraph}
\usepackage{CJKutf8}
\usepackage{amsfonts}
\usepackage{arydshln}
\usepackage{csquotes}
\usepackage{subfig}

\definecolor{TableShade}{gray}{0.96}
\definecolor{lightgrey}{RGB}{253, 253, 253}
\definecolor{newblue}{RGB}{218, 232, 252}
\definecolor{newgrey}{RGB}{240, 240, 240}

\newcommand{\reducedstrut}{\vrule width 0pt height .9\ht\strutbox depth .9\dp\strutbox\relax}
\newcommand{\TextShade}[1]{%
  \begingroup
  \setlength{\fboxsep}{0pt}%
  \colorbox{newgrey}{\reducedstrut#1\/}%
  \endgroup
}

\newcommand{\eg}{{e.g.}}
\newcommand{\ie}{{i.e.}}
\newcommand{\vs}{{vs. }}
\newcommand{\tc}[1]{\textcolor{darkblue}{#1}}
\newcommand{\dn}{{C$^3$}} %
\newcommand{\bc}{{B\textsubscript{c}}} %
\newcommand{\bn}{{B\textsubscript{n}}} %

\newcommand{\hhide}[1]{}

\newcommand{\ibf}[1]{\textbf{\emph{#1}}}

\newcommand\blfootnote[1]{%
  \begingroup
  \renewcommand\thefootnote{}\footnote{#1}%
  \addtocounter{footnote}{-1}%
  \endgroup
}

\usepackage{microtype}

\aclfinalcopy %
\title{Improving Machine Reading Comprehension with \\Contextualized Commonsense Knowledge}

\author{
  Kai Sun\textsuperscript{1}\textsuperscript{\dag} ~~ Dian Yu\textsuperscript{2}\textsuperscript{\dag} ~~ Jianshu Chen\textsuperscript{2} ~~ Dong Yu\textsuperscript{2} ~~ Claire Cardie\textsuperscript{1}\\
 \textsuperscript{1}Cornell University, Ithaca, NY \\
 \textsuperscript{2}Tencent AI Lab, Bellevue, WA \\
   ks985@cornell.edu, \{yudian, jianshuchen, dyu\}@tencent.com, cardie@cs.cornell.edu \\
}

\date{}

\begin{document}
\maketitle
\date{}

\blfootnote{$\dag$~Work was conducted when K. S. was an intern at the Tencent AI Lab, Bellevue, WA. Equal contribution.}

\begin{CJK*}{UTF8}{gkai}

\input{0_abstract}
%
\input{1_intro}           %

\input{2_extraction}
%

\input{3_teacher_student} %
\input{4_experiments}
%
\input{5_related_work}    %
\input{6_conclusion}      %

\bibliography{emnlp2020}
\bibliographystyle{acl_natbib}

\end{CJK*}
\end{document}

%% file: 0_abstract.tex
\begin{abstract}
In this paper, we aim to extract commonsense knowledge to improve machine reading comprehension. We propose to represent relations implicitly by situating structured knowledge in a context instead of relying on a pre-defined set of relations, and we call it contextualized knowledge. Each piece of contextualized knowledge consists of a pair of interrelated verbal and nonverbal messages extracted from a script and the scene in which they occur as context to implicitly represent the relation between the verbal and nonverbal messages, which are originally conveyed by different modalities within the script. We propose a two-stage fine-tuning strategy to use the large-scale weakly-labeled data based on a single type of contextualized knowledge and employ a teacher-student paradigm to inject multiple types of contextualized knowledge into a student machine reader. Experimental results demonstrate that our method outperforms a state-of-the-art baseline by a $4.3\%$ improvement in accuracy on the machine reading comprehension dataset {\dn}, wherein most of the questions require unstated prior knowledge.

\end{abstract}

%% file: 1_intro.tex
\section{Introduction}
\label{intro}

During the past few years, there is a trend of taking advantage of existing commonsense knowledge graphs such as ConceptNet~\cite{speer2017conceptnet} or automatically constructed graphs~\cite{zhang2020transomcs} to improve machine reading comprehension (MRC) tasks that contain a high percentage of questions requiring commonsense knowledge unstated in the given documents~\cite{mostafazadeh2016corpus,lai2017race,ostermann2018semeval,sundream2018,huang-etal-2019-cosmos}. In this paper, following the second line of work, we aim to extract commonsense knowledge from external unstructured corpora and explore using the structured knowledge to improve machine reading comprehension.

Typically, each piece of commonsense knowledge is represented as a triple that contains two phrases (\eg, (\emph{``finding a lost item''}, \emph{``happiness''}) and the relation (\eg, \textsc{Causes}) between phrases, which can be one of a small pre-defined set of relations~\cite{tandon2014webchild,speer2017conceptnet,sap2019atomic}. A carefully designed relation set is indispensable for many fundamental tasks such as knowledge graph construction. However, it is still unclear whether we need to explicitly represent relations if the final goal is to improve downstream tasks (\eg, machine reading comprehension) that do not directly depend on the reliability of relations in triples from other sources. Once we decide not to name relations, one natural question is whether we could implicitly represent relations between two phrases. We suggest that adding context in which the phrases occur may be useful as such a context constrains the possible relations between phrases without intervening in the relations explicitly~\cite{brezillon1998contextual}. Hereafter, we call a triple that contains a phrase pair and its associated context as a piece of \textbf{contextualized knowledge}.

\begin{table*}[!htb]
\centering
\footnotesize
\begin{tabular}{p{2.2cm}p{4cm}p{0.1cm}l}
\toprule
\rowcolor{lightgrey}
\textbf{scene} & \multicolumn{3}{l}{} \\
\midrule
\textbf{$\Box$}           & \multicolumn{3}{l}{Interior. Runaway office. Day.}    \\

Andy: & \multicolumn{3}{l}{I tried to ask her, but...}\\
Emily: & \multicolumn{3}{l}{You never ask Miranda. Anything. (\textbf{sighs}) All right, I’ll take care of the other stuff. You go to}\\
 & \multicolumn{3}{l}{Calvin Klein.}\\
Andy: & \multicolumn{3}{l}{Me?} \\
Emily: & \multicolumn{3}{l}{I’m sorry. Do you have a prior commitment? Is there some hideous pants convention?} \\
Andy:            & \multicolumn{3}{l}{So I just, what, go down to the Calvin Klein store and ask them...}    \\
\textbf{$\Diamond$}       & \multicolumn{3}{l}{\textbf{Emily rolls her eyes so hard they almost eject from her head.}}    \\
Emily:          & \multicolumn{3}{l}{You’re not going to the store.}                                       \\
Andy:            & \multicolumn{3}{l}{Of course not. \TextShade{I’m going...(\textbf{thinking})...to his house.}}          \\
Emily (\textbf{oh god}): & \multicolumn{3}{l}{You are catching on quickly. We always send assistants to a designer’s home on their very first} \\
& \multicolumn{3}{l}{day. You’re going to his showroom. I’ll give you the address.}\\
Andy:         & \multicolumn{3}{l}{Sorry. Got it. What’s the nearest subway stop?}    \\
Emily:          & \multicolumn{3}{l}{Good God. You do not. Under any circumstances. Take public transportation.}    \\
Andy:            & \multicolumn{3}{l}{I don’t?}    \\
\midrule
\rowcolor{lightgrey}
\textbf{type} & \textbf{nonverbal}   &       & \textbf{verbal}                   \\
\midrule
B\textsubscript{c}     &  oh god   &    & Emily: You are catching on \textelp{} I’ll give you the address.\\
I    &   sighs                          & & Emily: You never ask Miranda. Anything. All right  \textelp{} Klein.\\
I    &   thinking                          & & Andy: Of course not. I’m going......to his house.       \\
O    &   Emily rolls her eyes so hard     &    &  Andy: So I just, what, go down to the   \\
    &  they almost eject from her head.    &   & $~~~~~~~~~~~$Calvin Klein store and ask them...      \\
\bottomrule
\end{tabular}
\caption{A sample scene in a script and examples of verbal-nonverbal pairs extracted from this scene (all translated into English; \textelp{}: words omitted; $\Box$: scene heading; $\Diamond$: action line). The scene is regarded as the context of all the verbal-nonverbal pairs.}
\label{tab:sample}
\end{table*}

Besides verbal information that is written or spoken, it is well accepted that nonverbal information is also essential for face-to-face communication~\cite{jones2002research}. We regard related verbal and nonverbal information as the phrase pair; we treat the context in which the verbal-nonverbal pair occurs as the context. Such a triple can be regarded as a piece of commonsense knowledge as verbal and nonverbal information function together in communications, and this kind of knowledge is assumed to be known by most people without being formally taught. For example, as shown in Table~\ref{tab:sample}, the pause in \emph{``I’m going......to his house.''} is related to \emph{``thinking''}, the internal state of the speaker. We suggest film and television show scripts are good source corpora for extracting contextualized commonsense knowledge as they contain rich strongly interrelated verbal (\eg, utterances of speakers) and nonverbal information (\eg, body movements, vocal tones, or facial expressions of speakers), which is originally conveyed in different modalities within a short time period and can be easily separated from the scripts. Furthermore, a script usually contains multiple scenes, and the entire text of the scene from which the verbal-nonverbal pair is extracted can serve as the context. According to the relative position of a verbal-nonverbal pair in a scene, we use lexical patterns to extract four types of contextualized knowledge (Section~\ref{sec:extraction}).

To use contextualized knowledge to improve MRC, we randomly select nonverbal messages from the same script to convert each piece of knowledge into a weakly-labeled MRC instance (Section~\ref{sec:mrc_instance}). We propose a two-stage fine-tuning strategy to use the weakly-labeled MRC data: first, we train a model on the combination of the weakly-labeled data and the target MRC data that is human-annotated but relatively small-scale, and then, we fine-tune the resulting model on the target data alone (Section~\ref{sec:2ft}). We observe that training over the combination of all the data based on all types of contextualized knowledge does not lead to noticeable gains compared to using one type of knowledge. Therefore, we further use a teacher-student paradigm with multiple teacher models trained with different types of knowledge (Section~\ref{sec:ts}).

We evaluate our method on a multiple-choice MRC dataset \dn~\cite{sun2019investigating} in which most questions require prior knowledge such as commonsense knowledge besides the given contents. Experimental results demonstrate that our method leads to a $4.3\%$ improvement in accuracy over a state-of-the-art baseline~\cite{sun2019investigating,cui-2020-revisiting}. We also seek to transfer the knowledge to a different task by adapting the resulting student MRC model, which yields a $2.9\%$ improvement in F$1$ over a baseline on a dialogue-based relation extraction dataset DialogRE~\cite{yu-2020-dialogue}.

The main contributions are as follows: (\textbf{i}) we suggest that scripts can be a good resource for extracting contextualized commonsense knowledge, and our empirical results demonstrate the usefulness of contextualized knowledge for MRC tasks that require commonsense knowledge and the feasibility of implicitly representing relations by situating structured knowledge in a context; (\textbf{ii}) we propose a simple yet effective two-stage fine-tuning strategy to use large-scale weakly-labeled data; and (\textbf{iii}) we further show the effectiveness of a teacher-student paradigm to inject multiple types of contextualized knowledge into a single model.

\hhide{
The main contributions are as follows.
\begin{itemize}%
    \item we suggest that scripts can be a good resource for extracting contextualized commonsense knowledge, and our empirical results demonstrate the usefulness of contextualized knowledge for MRC tasks that require commonsense knowledge and the feasibility of implicitly representing relations by situating structured knowledge in a context.
    \item we propose a simple yet effective two-stage fine-tuning strategy to use large-scale weakly-labeled data.
    \item we further show the effectiveness of a teacher-student paradigm to inject multiple types of contextualized knowledge into a single model.
\end{itemize}
}

%% file: 2_extraction.tex
\section{Contextualized Knowledge Extraction}
\label{sec:extraction}

Both verbal information that is written or spoken and nonverbal information (\eg, body movements and facial expressions) are essential for face-to-face communication~\cite{jones2002research,calero2005power}. We propose to use interrelated verbal and nonverbal information as phrases in the traditional form of commonsense knowledge representation~\cite{speer2017conceptnet}.

We regard the interrelationship between such a verbal-nonverbal pair as a kind of commonsense knowledge because they function together in communications, and such knowledge is assumed to be known by most people without being formally told just as the definition of commonsense knowledge. We now introduce how to extract verbal-nonverbal pairs and extract the context in which it occurs. Formally we call a triple ($v$, $c$, $n$) as a piece of \textbf{contextualized knowledge}, containing a pair of related verbal information $v$ and nonverbal information $n$, as well as the associated context $c$. We choose to extract contextualized knowledge from film and television show scripts\footnote{As it is difficult to verify whether a text is written before a presentation (\ie, script) or during/after a presentation (\ie, transcript), we use \emph{scripts} throughout this paper.} as rich verbal and nonverbal messages frequently co-occur in scripts, and they can be easily separated.
Scenes in a script are separated by blank lines. According to the relative position of verbal and nonverbal information, we extract four types of contextualized knowledge (B\textsubscript{c}, B\textsubscript{n}, I, and O) as follows.

\begin{itemize}
    \item Beginning: the nonverbal information $n$ appears after a speaker name and before the speaker's utterance. We regard the speaker name and the corresponding utterance as $v$.
    \begin{itemize}
        \item Clean (B\textsubscript{c}): We only extract nonverbal information $n$ within parentheses.
        \item Noisy (B\textsubscript{n}): The first span of a turn, followed by a colon, can also contain both a speaker name and nonverbal information about this speaker. It usually happens when a script is written without strictly following a standard screenplay format. We remove the phrase that is a potential speaker name from the span and regard the remaining text in the span as $n$. We roughly regard a phrase as a speaker name if it appears in the first span of other turns in the same scene. %
    \end{itemize}
    \item Inside (I): We only extract nonverbal information $n$ enclosed in parentheses, which appears within an utterance. All the information in the same turn except $n$ is treated as $v$.
    \item Outside (O): Here $n$ is an action line that mainly describes what can be seen or heard by the audience, marked by $\Diamond$ in Table~\ref{tab:sample}. We regard the turn (if it exists) before the action line as its corresponding $v$.
\end{itemize}

We do not extract phrases in parentheses or action lines as nonverbal information if they are terminologies for script writing such as \emph{``O.S."}, \emph{``CONT’D''}, \emph{``beat''}, \emph{``jump cut"}, and \emph{``fade in''}.\footnote{We will release the stop word list along with the code.} All types of contextualized knowledge extracted from a scene share the same context $c$, \ie, the scene itself. We do not exploit the scene heading mostly about when and where a scene takes place (marked by $\Box$ in Table~\ref{tab:sample}), as it is intentionally designed to cover the content of the whole scene, which is already used as context.

\section{Instance Generation}
\label{sec:mrc_instance}

As most current MRC tasks requiring commonsense knowledge are usually in a multiple-choice form, we mainly discuss how to convert the extracted triples into multiple-choice instances and leave its extension to other types (\eg, extractive or abstractive) of MRC tasks for future research. 

We generate instances for each type of contextualized knowledge. For each triple ($v$, $c$, $n$), we remove $n$ from context $c$, and we regard the remaining content as the reference document, verbal information $v$ as the question, and the nonverbal information $n$ as the correct answer option. To generate distractors (\ie, wrong answer options), we randomly select $N$ items from all the unique nonverbal information in other triples, which belong to the same type of contextualized knowledge and are extracted from the same script as ($v$, $c$, $n$). Note that we only generate one instance based on each triple, while it is easy to generate more instances by changing distractors.

%% file: 3_teacher_student.tex
\begin{figure*}[h!]
   \begin{center}
   \includegraphics[width=0.76\textwidth]{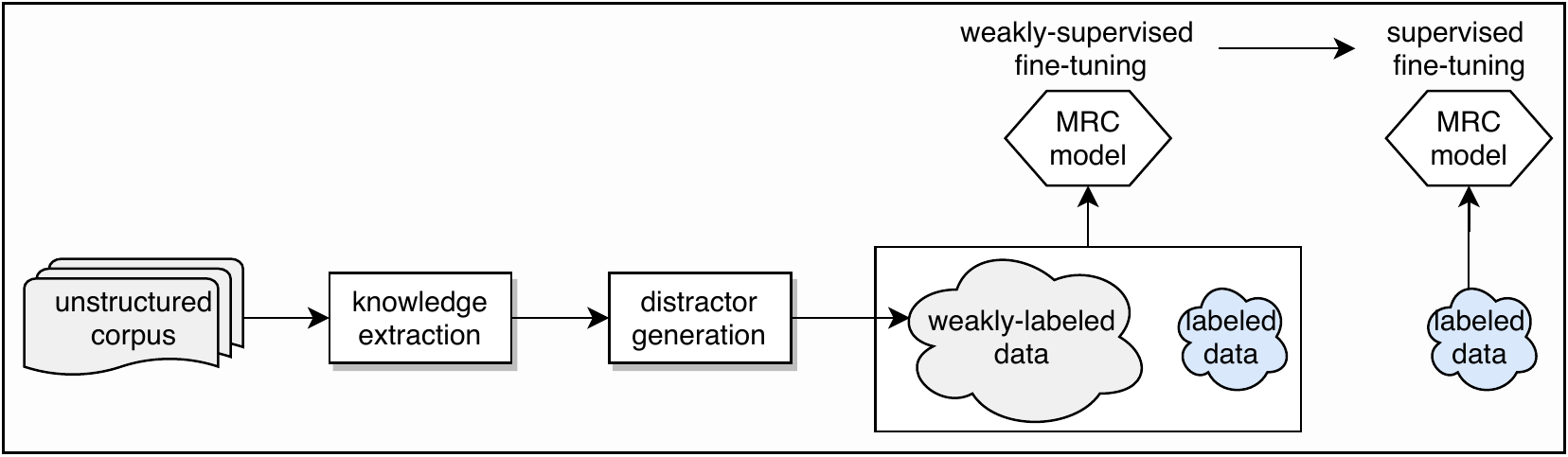}
   \end{center}
 \caption{Two-stage fine-tuning framework overview (one type of contextualized knowledge is involved).}
 \label{fig:overview_single}
\end{figure*}

\section{Two-Stage Fine-Tuning}
\label{sec:2ft}
As mentioned previously, we aim to use the constructed weakly-labeled data to improve a downstream MRC task ({\dn} in this paper). Given weakly-labeled data generated based on \textbf{one} type of contextualized knowledge (\eg, {\bc} or I) extracted from scripts, we first use the weakly-labeled data in conjunction with the training set of {\dn} as the training data to train the model and then fine-tune the resulting model on {\dn} as illustrated in Figure~\ref{fig:overview_single}. We do not adjust the ratio of clean data to weakly-labeled data observed during training as previous joint training work on other tasks such as machine translation~\cite{edunov-etal-2018-understanding}.

Another way is to perform separate training: we first train the model on the weakly-labeled data and then fine-tune it on {\dn}. In our preliminary experiment, we observe that joint training leads to better performance, and therefore we apply it in all the experiments. See performance comparisons of joint and separate training in Section~\ref{sec:ablation}.

\begin{figure*}[h!]
   \begin{center}
   \includegraphics[width=0.98\textwidth]{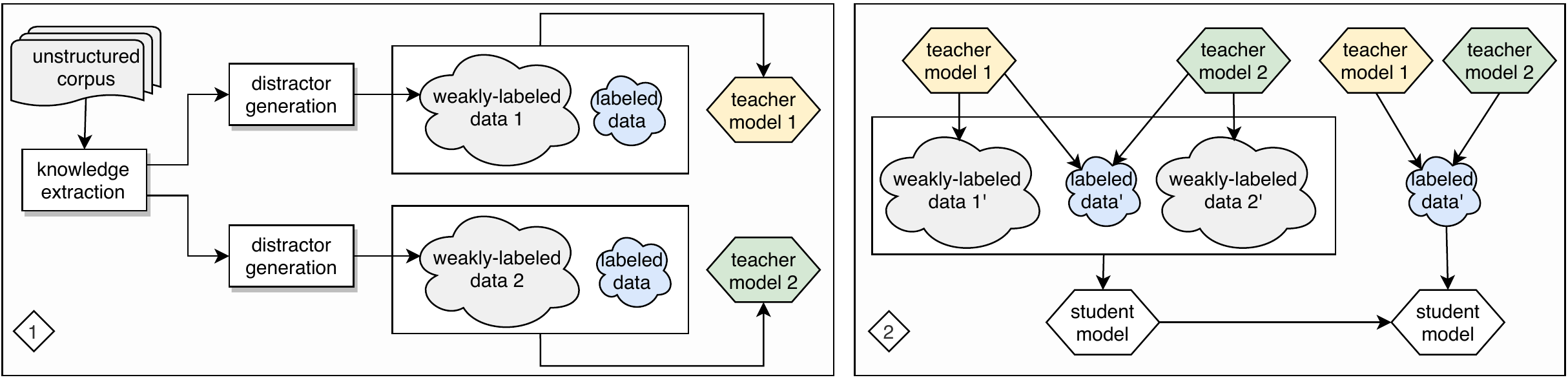}
   \end{center}
 \caption{Teacher-student paradigm overview (multiple types of contextualized knowledge are involved). To save space, we only show the case that involves two types of contextualized knowledge. }
 \label{fig:ts}
\end{figure*}

\section{Teacher-Student Paradigm}
\label{sec:ts}

As introduced in Section~\ref{sec:mrc_instance}, we have \textbf{multiple} sets of weakly-labeled data, each corresponding to one type of contextualized knowledge. We observe that simply combining all the data, either in joint training or separate training, does not lead to noticeable gains compared to using one type of contextualized knowledge. Inspired by the previous work~\cite{you2019teach} that trains a student automatic speech recognition model with multiple teacher models, and each teacher model is trained on a domain-specific subset with a unique speaking style, we employ a teacher-student paradigm to inject multiple types of contextualized knowledge into a single student machine reader.

Let $V$ denote a set of labeled instances, $W_1, \dots, W_\ell$ denote $\ell$ sets of weakly-labeled instances, and $W = \bigcup_{1 \le i \le \ell}W_i$. 
For each instance $t$, we let $m_t$ denote its total number of answer options, and ${\ibf h}^{(t)}$ be a hard label vector (one-hot) such that ${\ibf h}^{(t)}_j = 1$ if the $j$-th option is labeled as correct. We train $\ell$ teacher models, denoted by $\mathcal{T}_1, \dots, \mathcal{T}_\ell$, and optimize $\mathcal{T}_i$ by minimizing $\sum_{t \in V\cup W_i} L_1(t, \theta_{\mathcal{T}_i})$. $L_1$ is defined as 
\begin{equation*}
L_1(t, \theta) = - \sum_{1 \le k \le m_t} {\ibf h}^{(t)}_k ~ \log p_\theta(k\,|\,t), 
\end{equation*}
where $p_\theta(k\,|\,t)$ denotes the probability that the $k$-th option of instance $t$ is correct, estimated by the  model with parameters $\theta$.

We define soft label vector ${\ibf s}^{(t)}$ such that
\begin{equation*}\small
    {\ibf s}^{(t)}_k =
    \begin{cases}
     \lambda~{\ibf h}^{(t)}_k + (1 - \lambda) \displaystyle \sum_{1\le j \le \ell}\frac{1}{\ell} p_{\theta_{\mathcal T_j}}(k\,|\,t)& t\in V \\
     \lambda~{\ibf h}^{(t)}_k + (1 - \lambda) p_{\theta_{\mathcal T_i}}(k\,|\,t)& t\in W_i \\
    \end{cases},
\end{equation*}
where $\lambda\in [0, 1]$ is a weight parameter, and $k = 1,\dots, m_t$. 

We then train a student model, denoted by $\mathcal S$, in a two-stage fashion. In stage one (\ie, weakly-supervised fine-tuning), we optimize $\mathcal S$ by minimizing $\sum_{t \in V\cup W}L_2(t, \theta_{\mathcal S})$, where $L_2$ is defined as
\begin{equation*}
    L_2(t, \theta) = -\sum_{1\le k\le m_t} {\ibf s}^{(t)}_k ~ \log p_{\theta}(k\,|\,t).
\end{equation*}
In stage two (\ie, supervised fine-tuning), we further fine-tune the resulting $\mathcal S$ after stage one by minimizing $\sum_{t\in V} L_2(t, \theta_{\mathcal S})$. See Figure~\ref{fig:ts} for an overview of the paradigm.

%% file: 4_experiments.tex
\section{Experiment}

\subsection{Data}

We collect 8,166 scripts in Chinese, and most of them are intended for films and television shows.\footnote{https://www.1bianju.com.} After segmentation and filtering, we obtain 199,280 scenes, each of which contains at least one piece of contextualized knowledge defined in Section~\ref{sec:extraction}. We generate four sets of weakly-labeled data based on the scenes. For comparison, we also use existing human-annotated triples about commonsense knowledge in the Chinese version of ConceptNet~\cite{speer2017conceptnet}. We set the number of distractors $N$ (Section~\ref{sec:mrc_instance}) to five when we convert structured triples into MRC instances.

For evaluation, we use \dn, a free-form multiple-choice MRC data for Chinese collected from Chinese-as-a-second-language exams~\cite{sun2019investigating}. About $86.8\%$ of questions in \dn~involve prior knowledge (\ie, linguistic, domain-specific, and commonsense knowledge) not provided in the given texts, and all instances are carefully designed by experts such as second-language teachers. Each instance consists of a document, a question, and multiple answer options; only one answer option is correct. See Table~\ref{tab:statistics} for data statistics. 

While we focus on scripts and datasets in Chinese in this study, our extraction and training methods are not limited to a particular language.

\begin{table}[ht!]
\centering
\small
\begin{tabular}{lll}
\toprule
data & type of construction                      & \# of instances \\
\midrule
\dn   & human-annotated             & 19,577  \\
\midrule
ConceptNet  & human-annotated            & 737,534  \\
\midrule
\bc   & weakly-labeled             & 105,622  \\
\bn   & weakly-labeled             & 198,053  \\
I     & weakly-labeled             & 204,750  \\
O     & weakly-labeled             & 192,391  \\
\bc + \bn + I + O &  weakly-labeled      & 700,816  \\
\bottomrule
\end{tabular}
\caption{\label{tab:statistics} Data statistics.}
\end{table}

\begin{table*}[h!]
\centering
\small
\begin{tabular}{llcclccc}
\toprule

index   & \multicolumn{2}{c}{weakly-supervised fine-tuning} & &  \multicolumn{2}{c}{supervised fine-tuning} & dev & test \\
\cline{2-3} \cline{5-6}
        & data                  & teacher-student   &  & data     & teacher-student        &  &      \\
   \midrule
\tc 0                               &  --   & --   & & \dn   &  --  & 73.9   & 73.4 \\
\midrule
\rowcolor{TableShade}
\tc 1  &  \dn + \bc                 & -- & & --     &  --  & 71.1   & 71.7 \\
\tc 2  &  \dn + \bc                 & -- & & \dn   & -- & 74.5   & 74.0 \\
\rowcolor{TableShade}
\tc 3  &  \dn + \bn                 & --  & & --    & -- & 71.3   & 72.0 \\
\tc 4  &  \dn + \bn                 & -- & & \dn   & -- & 74.6   & 74.5 \\
\rowcolor{TableShade}
\tc 5  &  \dn + I                   & --&  & --     & -- & 73.5   & 72.8 \\
\tc{6}  &  \dn + I                  & -- & & \dn   & -- & \textbf{75.6}  & \textbf{74.9} \\
\rowcolor{TableShade}
\tc{7}  &  \dn + O                  & -- & & --     & -- & 72.4   & 72.7 \\  
\tc{8}  &  \dn + O                  & -- & & \dn   & -- & 75.4   & 74.9 \\ 
\midrule
\rowcolor{TableShade}
\tc{9}  &   \dn + \bc + \bn + I + O  & -- & & --  & -- & 71.6   & 71.0 \\
\tc{10}  &  \dn + \bc + \bn + I + O & --  & & \dn    & -- & 75.6   & 75.2 \\
\tc{11}  &  \dn + \bc + \bn + I + O & $\checkmark$ & & \dn  & -- & 76.5    & 76.4 \\
\tc{12}  &  \dn + \bc + \bn + I + O & $\checkmark$ & & \dn  & $\checkmark$ & \textbf{77.4}  & \textbf{77.7}\\
\bottomrule
\end{tabular}
\caption{\label{tab:performance} Average accuracy (\%) on the development and test sets of the {\dn} dataset.}
\end{table*}

\subsection{Implementation Details}
 In our experiments, we follow \citet{sun2019investigating} for the model architecture consisting of a pre-trained language model and a classification layer on top of the model. We use RoBERTa-wwm-ext-large~\cite{cui-2020-revisiting} as the pre-trained language model, which achieves state-of-the-art performance on {\dn} and many other natural language understanding tasks in Chinese~\cite{xu2020clue}. We leave the exploration of more pre-trained language models for future work. When the input sequence length exceeds the limit, we repeatedly discard the last turn in the context, or the first turn if the last turn includes the extracted verbal information. We train a model for one epoch during the weakly-supervised fine-tuning stage and eight epochs during the supervised fine-tuning stage. We set $\lambda$ (defined in Section~\ref{sec:ts}) to $0.5$ in all experiments based on the rationale that we can make best use of the soft labels while at the same time making sure $\arg\max_k{\ibf s}^{(t)}_k$ is always the index of the correct answer option for instance $t$. Carefully tuning $\lambda$ on the development set may lead to further improvements, which is not the primary focus of this paper.

\subsection{Main Results and Discussions}
\label{sec:baseline}

Table~\ref{tab:performance} reports the main results. The baseline accuracy ($73.4\%$ \{\tc{0}\}) is slightly lower than previously reported using the same language model\footnote{https://github.com/CLUEbenchmark/CLUE.} as we report the average accuracy over five runs with different random seeds for all our supervised fine-tuning results. For easy reference, we indicate the index for each result in curly brackets in the following discussion. Obviously, the performance of a model after the first fine-tuning stage over the combination of the \dn~dataset and much larger weakly-labeled data is worse (\eg, $71.7\%$ \{\tc{1}\}) than baseline performance (\{\tc{0}\}). Further fine-tuning the resulting model on the {\dn} dataset consistently leads to improvements (\eg, $74.0\%$ \{\tc{2}\} and $74.5\%$ \{\tc{4}\}) over the baseline \{\tc{0}\}, demonstrating the effectiveness of the \textbf{two-stage} fine-tuning strategy for using large-scale weakly-labeled data. We will discuss the critical role of the target task's data (\ie, {\dn}) in the weakly-supervised fine-tuning stage in the next subsection. Following this strategy, \textbf{each} of the weakly-labeled data based on one type of contextualized knowledge can boost the final performance (\{\tc{2, 4, 6, 8}\}); the magnitude of accuracy improvement is $1.2\%$ on average. 

When we combine all the weakly-labeled data in the first fine-tuning stage, the performance gain after the second round of fine-tuning ($75.2\%$ \{\tc{10}\}) is not as impressive as expected, given the best performance achieved by only using one set ($74.9\%$ \{\tc{6}\}). As a comparison, our \textbf{teacher-student paradigm} that trains multiple teacher models with different types of weakly-labeled data leads to up to $3.7\%$ improvement in accuracy (\{\tc{12}\} \vs \{\tc{2, 4, 6, 8}\}). The advantage is reduced but still exists even when we use the original hard labels instead of soft labels in the second fine-tuning stage ($76.4\%$ \{\tc{11}\}).

\subsection{Ablation Studies and Discussions}
\label{sec:ablation}
We have shown that our proposed teacher-student paradigm helps inject multiple types of knowledge into the baseline. We conduct ablation studies to examine critical factors. We first remove the context (\ie, scene) from each instance in the weakly-labeled data and leave it empty. All other aspects of this baseline remain the same as \{\tc{12}\} in Table~\ref{tab:performance}. We also remove the \dn~dataset from the weakly-supervised fine-tuning stage when we train teacher and student models (Figure~\ref{fig:ts}) and only use \dn~during the supervised fine-tuning stage. We observe that accuracy decreases in both conditions (Table~\ref{tab:ablation}), demonstrating the usefulness of contexts in contextualized knowledge for improving machine reading comprehension and the importance of involving the human-annotated data of the target task, although small-scale, in the weakly-supervised fine-tuning stage.

\begin{table}[ht!]
\centering
\small
\begin{tabular}{lll}
\toprule
method                                                    & dev   & test \\
\midrule
\{\tc{12}\} in Table~\ref{tab:performance}                & 77.4  & 77.7  \\
\{\tc{12}\} w/o context in weakly-labeled data            & 76.8  & 76.6  \\
\{\tc{12}\} w/o using \dn~in the 1st FT          & 76.6  & 76.2 \\
\bottomrule
\end{tabular}
\caption{\label{tab:ablation} Ablation results on the development and test sets of the \dn~dataset (FT: fine-tuning).}
\end{table}

\begin{table}[ht!]
\centering
\small
\begin{tabular}{lccc}
\toprule
category                        & \{\tc{0}\} & \{\tc{12}\}  & $\Delta$ \\
\midrule
Matching                        & 90.0    & \textbf{94.7} & 4.7   \\
Prior Knowledge                 & 69.5    & \textbf{75.3} & 5.8   \\ 
$\diamond$ Linguistic           & 73.8    & \textbf{77.8} & 4.0  \\
$\diamond$ Commonsense        & 68.0    & \textbf{74.4} & 6.4   \\
$\diamond$ Domain-specific$^\star$  & 13.3    & \textbf{20.0} & 6.7 \\
\bottomrule
\end{tabular}
\caption{\label{tab:dev}Average accuracy (\%) on the annotated development set of \dn~per question category ($\star$: the domain-specific category only contains three instances).}
\end{table}

\begin{table*}[ht!]
\centering
\small
\begin{tabular}{lllllll}
\toprule
notes & \multicolumn{4}{c}{weakly-labeled data}  & dev  & test \\
\cline{2-5}
 & structured knowledge   & document      & question  & answer    &   &  \\
\midrule
\{\tc{0}\} in Table~\ref{tab:performance}  & -- & --    & --   & -- & 73.9  & 73.4     \\
\midrule
\{\tc{10}\} in Table~\ref{tab:performance}  & contextualized knowledge & scene    & verbal   & nonverbal & 75.6   & 75.2     \\
\{\tc{10}\} w/o context &  contextualized knowledge & empty    & verbal   & nonverbal  & 74.9  & 74.2      \\
i   &  ConceptNet               & empty    & subject   & object    & 74.0    & 72.7    \\
ii  &  ConceptNet               & relation type & subject   & object    & 74.6    & 74.1      \\

\bottomrule
\end{tabular}
\caption{\label{tab:conceptnet} Average accuracy (\%) on the development and test sets of the \dn~dataset using weakly-labeled data constructed based on contextualized knowledge or ConceptNet.}
\end{table*}
As we may require one or multiple types of prior knowledge to answer a question, we study the impacts of the contextualized knowledge on different types of questions based on the annotated subset ($300$ instances) released along with the dataset. As shown in Table~\ref{tab:dev}, our method generally improves performance on all types of questions, especially those that require commonsense knowledge.

Considering the similarity of B\textsubscript{c} and B\textsubscript{n} in the relative position of verbal and nonverbal information in a scene, we also experiment by merging B\textsubscript{c} and B\textsubscript{n} into a single set and then training three teacher models instead of four used for training the student model. Results show that it achieves a similar accuracy ($77.5\%$) to the four-teacher setting (\{\tc{12}\} in Table~\ref{tab:performance}). For further improvement, it may be a promising direction to train more teachers with diverse types or forms of external knowledge.

\subsection{A Comparison Between Contextualized Knowledge and ConceptNet}

Most of the existing commonsense knowledge graphs are in English. Therefore, we only compare contextualized knowledge with the Chinese version of a human-annotated commonsense knowledge graph ConceptNet. Each triple in ConceptNet is represented as (subject, relation type, object) (\eg, (\emph{``drink water''}, \textsc{causes}, \emph{``not thirsty''})). We experiment with three types of input sequences when we convert triples into MRC instances: (i) leave the document empty in each instance and (ii) use the relation type as the document. We randomly select phrases in ConceptNet other than the phrases in each triple as distractors.

For a fair comparison, we compare (ii) with baseline \{\tc{10}\} in Table~\ref{tab:performance} as it follows the same two-stage fine-tuning without using the teacher-student paradigm. To compare with (i), we run an ablation test of \{\tc{10}\} by removing contexts from weakly-labeled MRC instances. The amounts of weakly-labeled instances based on contextualized knowledge and ConceptNet are similar (Table~\ref{tab:statistics}). The results in Table~\ref{tab:conceptnet} reveal that under the two-stage fine-tuning framework, introducing ConceptNet yields up to $0.7\%$ in accuracy, but using contextualized knowledge gives a bigger gain of $1.8\%$ in accuracy. Furthermore, removing contexts from weakly-labeled instances hurts performance, consistent with our observation in Section~\ref{sec:ablation}.

We do not dismiss the construction and use of commonsense knowledge with a well-defined schema and admit that the form of contextualized knowledge representation is not concise enough for easy alignment with existing commonsense knowledge graphs or knowledge graph completion. However, we argue that contexts can tacitly state the relation between phrases, and this kind of commonsense knowledge is helpful for MRC.

\subsection{Transferring to Relation Extraction}

Seeking to transfer knowledge to other tasks, we take a relation extraction task DialogRE~\cite{yu-2020-dialogue} as a case study. The task aims to predict relations between an argument pair based on a given dialogue. We replace the classification layer of an MRC model with a multi-class multi-label classification layer following the baseline released by~\newcite{yu-2020-dialogue} and fine-tune the whole architecture on the Chinese version of the DialogRE dataset.

We compare the performance of methods that use different weights for parameter initialization except for the classification layer, which is randomly initialized. As shown in Table~\ref{tab:dialogre}, we achieve an improvement of $2.9\%$ in F$1$ and $3.1\%$ in $\text{F}1_\text{c}$ by adapting our best-performing machine reading comprehension model. The metric $\text{F}1_\text{c}$ is used to encourage a model to identify relations between arguments as early as possible rather than after reading the whole dialogue. Introducing \dn~alone also allows us to achieve a slight gain over the baseline. It might be interesting to study the relevance between document/dialogue-based relation extraction and machine reading comprehension to boost the performance of the two types of tasks.

\begin{table}[h!]
\centering
\small
\begin{tabular}{lllll}
\toprule
\multirow{2}{*}{parameter initialization}         & \multicolumn{2}{c}{dev} & \multicolumn{2}{c}{test} \\
                          & F$1$    & $\text{F}1_\text{c}$      &  F$1$         & $\text{F}1_\text{c}$         \\
\midrule
RoBERTa-wwm-ext-large                         & 64.9     & 60.3         &  64.4    &   59.2         \\
\{\tc{0}\} in Table~\ref{tab:performance}                       &  66.4        &  61.6      &   65.0   &   60.3         \\
\{\tc{12}\} in Table~\ref{tab:performance}     & \textbf{67.1}   & \textbf{62.9}         &  \textbf{67.3}    &   \textbf{62.3}         \\
\bottomrule
\end{tabular}
\caption{\label{tab:dialogre}Average F$1$ (\%) and $\text{F}1_\text{c}$ (\%) on the DialogRE dataset.}
\end{table}

%% file: 5_related_work.tex
\section{Related Work}

\subsection{Contextualized Knowledge}
We mainly discuss the external contextualized knowledge that is not directly relevant with a target task as retrieving relevant pieces of evidence from an external source for instances of a target task is not the focus of this paper. A common solution to obtain external contextualized knowledge is to utilize existing knowledge bases via distant supervision. For example,~\newcite{ye2019align} align triples in ConceptNet~\cite{speer2017conceptnet} with sentences from Wikipedia. We extract contextualized knowledge from scripts, where contexts (\ie, scenes) are naturally aligned with verbal-nonverbal pairs to avoid noise from distant supervision.

Our work is also related to commonsense knowledge extraction, which relies on human-annotated triples~\cite{xu-etal-2018-automatic,bosselut2019comet}, high-precision syntactic or semantic patterns~\cite{zhang2020transomcs,zhou-etal-2020-temporal} specific to each relation, or existing lexical databases ~\cite{tandon2014webchild,tandon2015knowlywood}. In comparison, we skip the step of offering a name of the relation between two phrases and focus on extracting structured knowledge in its context. Our language-independent knowledge extraction does not require any training data and does not rely on a high-quality semantic lexicon or a syntactic parser, which is usually unavailable for many non-English languages. 

\subsection{Weak Supervision for Machine Reading Comprehension}

As it is expensive and time-consuming to crowdsource or collect a large-scale, high-quality dataset, weak supervision has received much attention throughout the history of machine reading comprehension. Various forms of weak supervision are studied, mostly based on existing resources such as pre-trained semantic/syntactic parsers~\cite{smith-etal-2015-strong,wang2015machine,liu-etal-2017-structural} or natural language inference systems~\cite{pujari-goldwasser-2019-using,wang2019evidence}, knowledge bases~\cite{wang2018yuanfudao,wang2019explicit,yang-2019-enhancing-pre}, and linguistic lexicons~\cite{sun2019probing}. Compared to previous work, we focus on generating large-scale weakly-labeled data using the contextualized knowledge automatically extracted from unstructured corpora.

\subsection{Semi-Supervised Learning for Machine Reading Comprehension}

Previous semi-supervised methods that leverage internal or external unlabeled texts usually generate question and answer based on the content of the same sentence~\cite{yang-etal-2017-semi,wang-etal-2018-multi-perspective,dhingra-etal-2018-simple}. Besides the unlabeled texts, previous studies~\cite{yuan2017machine,yu2018qanet,zhang-bansal-2019-addressing,zhu-2019-learning,dong2019unified,alberti-etal-2019-synthetic,asai-hajishirzi-2020-logic} also heavily rely on the labeled instances of the target machine reading comprehension task for data augmentation. In comparison, we focus on generating non-extractive instances without using any task-specific patterns or labeled data, aiming to improve machine reading comprhension tasks that require substantial prior knowledge such as commonsense knowledge.

Another line of work develops unsupervised approaches~\cite{lewis-etal-2019-unsupervised,li-etal-2020-harvesting,fabbri-etal-2020-template} for extractive machine reading comprehension tasks. However, there is still a large performance gap between unsupervised and state-of-the-art supervised methods.

\subsection{Knowledge Integration}

Our teacher-student paradigm for knowledge integration is most related to multi-domain teacher-student training for automatic speech recognition~\cite{you2019teach} and machine translation~\cite{wang2019go}. Instead of clean domain-specific human-labeled data, each of our teacher models is trained with weakly-labeled data. Due to the introduction of large amounts of weakly-labeled data, the data of the target machine reading comprehension task (with hard or soft labels) is used during all the fine-tuning stages of both teacher and student models.

%% file: 6_conclusion.tex
\section{Conclusions}

In this paper, we aim to extract contextualized commonsense knowledge to improve machine reading comprehension. We propose to situate structured knowledge in a context to implicitly represent the relationship between phrases, instead of relying on a pre-defined set of relations. We extract contextualized knowledge from film and television show scripts as interrelated verbal and nonverbal messages frequently co-occur in scripts. We propose a two-stage fine-tuning strategy to use the large-scale weakly-labeled data and employ a teacher-student paradigm to inject multiple types of contextualized knowledge into a single student model. Experimental results demonstrate that our method outperforms a state-of-the-art baseline by a $4.3\%$ improvement in accuracy on the multiple-choice machine reading comprehension dataset {\dn}, wherein most of the questions require unstated prior knowledge, especially commonsense knowledge. %